\documentclass[journal]{IEEEtran}
\ifCLASSINFOpdf
\else
\fi

\usepackage{stfloats}
\usepackage[noend]{algpseudocode}
\usepackage{algorithmicx, algorithm}
\usepackage{graphicx}
\usepackage{float}
\usepackage{amsmath}
\begin{document}
%
\title{Knowing the Past to Predict the Future: Reinforcement Virtual Learning}

%
%
%

\author{Peng~Zhang*,
        Yawen~Huang*$^{\S}$,
        Bingzhang~Hu,
        Shizheng~Wang, Haoran Duan\\
        Noura Al Moubayed,
        Yefeng Zheng,~\IEEEmembership{Fellow,~IEEE},
        and~Yang~Long$^{\S}$,~\IEEEmembership{Member,~IEEE}
\thanks{* Equal contribution.}
\thanks{$^{\dag}$ Corresponding authors: Dr Yong Long (e-mail: yang.long@durham.ac.uk) and Dr Yawen Huang (e-mail: yawenhuang@tencent.com).}
\thanks{Dr Yawen Huang and Dr Yefeng Zheng are with Tencent Jarvis Lab, Shenzhen, China. Email: yawenhuang@tencent.com, yefengzheng@tencent.com}
\thanks{Peng Zhang, Haoran Duan, Dr Noura Al Moubayed and Dr Yong Long are with the Department of Computer Science, Durham University, UK. Email:peng.zhang@durham.ac.uk, haoran.duan@ieee.org, noura.al-moubayed@durham.ac.uk, yang.long@ieee.org}
\thanks{Dr Bingzhang Hu is with Hefei Institutes of Physical Science, Chinese Academy of Sciences, China. Email: hubzh@aiofm.ac.cn}
\thanks{Dr Shizheng Wang is with Institute of Microelectronics of the Chinese Academy of Sciences, Beijing, China. Email: shizheng.wang@foxmail.com}
\thanks{This work is also supported by The UK MRC Innovation Fellowship with ref MR/S003916/1.}}

%
%

\markboth{}%
{Zhang \MakeLowercase{\textit{\textit{et al.}}}: Knowing the Past to Predict the Future: Reinforcement Virtual Learning}
%



\maketitle

\begin{abstract}
\textit{Reinforcement Learning} (RL)-based control system has received considerable attention in recent decades. However, in many real-world problems, such as \textit{Batch Process Control}, the environment is uncertain, which requires expensive interaction to acquire the state and reward values. In this paper, we present a cost-efficient framework, such that the RL model can evolve for itself in a \textit{Virtual Space} using the predictive models with only historical data. The proposed framework enables a step-by-step RL model to predict the future state and select optimal actions for long-sight decisions. The main focuses are summarized as: 1) how to balance the long-sight and short-sight rewards with an optimal strategy; 2) how to make the virtual model interacting with real environment to converge to a final learning policy. Under the experimental settings of \textit{Fed-Batch Process}, our method consistently outperforms the existing state-of-the-art methods.
\end{abstract}



%
\IEEEpeerreviewmaketitle

\section{Introduction}
\IEEEPARstart{B}{atch} processes, as an important chemical process, are expected to generate higher value products, such as desirable chemicals, polymers and pharmaceuticals \cite{ccta}, which have received considerable attention in recent years. 
Due to the rapid evolution of diversely customised chemical processes, fed-batch is then considered to be one of the most popular approaches of responsive manufacturing. Among the fed-batch and batch process operations, the maximum end-of-batch product quality is the most noteworthy \cite{ccta}. Batch processes usually face a dilemma in optimisation and control treatment, due to the rapid time-varying characteristics, non-steady operations and non-linearity batch polymerisation reactors \cite{mm}.

The existing solutions are sought from \textit{Modern Control Theory}, which was experienced a rapid improvement on their optimisation mechanism. A number of optimal control approaches, \textit{e.g.}, Proportional-Integral(PI), Proportional–Integral–Derivative(PID) and fuzzy control, have been applied in various disciplines. For example, Khalili \textit{et al.} \cite{kha} proposed an optimal sliding mode control in biology. Trajectory optimization was then presented and applied in robotics by Carius \textit{et al.} \cite{ca}. Wei \textit{et al.} \cite{wei} applied such an optimal control to operate and optimize motor. Optimal control was applied in fractional order dynamical systems by Mohammadzadeh \textit{et al.} \cite{mo} and Razminia \textit{et al.} \cite{ra}. Das \textit{et al.} \cite{das} and Bian \textit{et al.} \cite{bian} applied optimal control in power systems. In parallel, many efforts were paid in chemical engineering by Shi \textit{et al.} \cite{shi}, Cui \textit{et al.} \cite{cui} and Sun \textit{et al.} \cite{sun}.
\begin{figure*}[h]
\centering
\includegraphics[width=8cm, height=3.5cm]{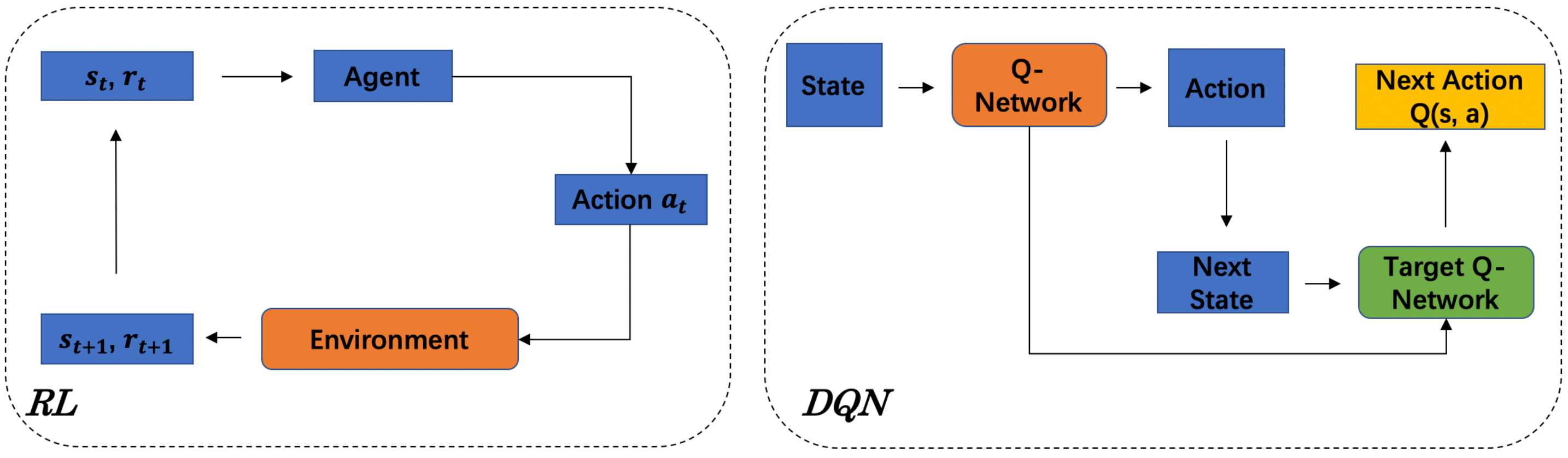}
\includegraphics[width=10cm, height=3.5cm]{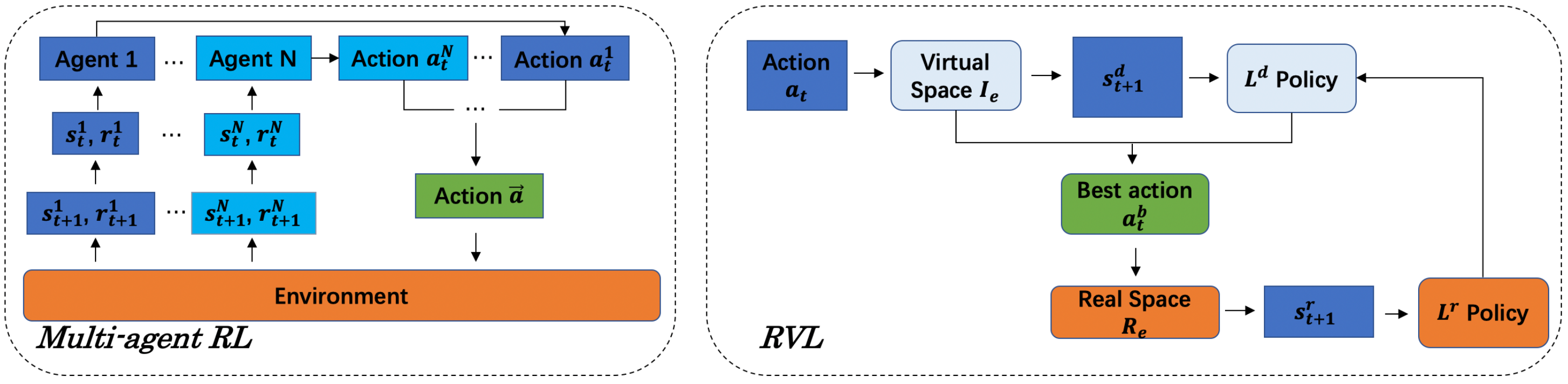}
\caption{Several widely used methods of reinforcement learning, such as the traditional RL, DQN and multi-agent reinforcement. The differences between the proposed RVL and the existing models are highlighted.}
\end{figure*}


With the rapid development of \textit{Machine Learning} technologies, an emerging trend of modern control systems has been introduced by expolring the advanced data-driven strategies, \textit{e.g.}, neural networks and hybrid computational intelligence algorithms \cite{ccta}. Particularly, \textit{Reinforcement Learning} (RL) models have manifested the application values in many fields, such as computer vision \cite{yang} \cite{bing}, games \cite{human} and medicine \cite{gao}. With the development of neural networks, Deep Learning (DL) and RL models have been successfully applied in various chemical processes. For example, Jie \textit{et al.} \cite{jie} applied the recurrent neural fuzzy network in fed-batch process. Shah and Gopal \cite{8} applied Q-learning to solve the problem of adaptive control of a nonlinear chemical process. The operation of robots was optimized and controlled by Said and Guido \cite{9}. Lambe \textit{et al.} \cite{10} utilized Q-learning to address the path length of Nanobots. Spielberg \textit{et al.} \cite{11} further leveraged Q-learning for processing control. In addition, an improved stochastic multi-step action Q-learning algorithm (SMSA) \cite{ccta} and a modified multi-step action Q-learning algorithm (MMSA) \cite{mm} were proposed to handle both control and optimization of fed-batch processes. Although RL has been applied in different chemical processes, it still lacks exploration in fed-batch processes.

In this paper, a new structure of Reinforcement Virtual Learning (RVL) is proposed to control and optimize the fed-batch process. The novelty can be summed up as that the virtual space is explored and cooperated with RL, which means that a virtual environment can be predicted and created by previous data and the RL agent can further interact with virtual environment to learn. Specifically, a simple and conventional prediction model is explored combined with RL to generate a more effective and flexible method. We summarize our main contributions below:
\begin{itemize}
\item The learned agent of RL through interaction with virtual environment can acquire a virtual learning policy. When the agent of RL interacts with real environment, this virtual learning policy can introduce and modify the agent to learn a real learning policy. According to this cooperation, RL can control and optimize the process in an uncertain environment. Also, previous historical information can be utilized adequately.
\item Besides previous historical information, the proposed RVL can leverage future information as well. In terms of the virtual environment and learning policy, the results of future approximation control can be obtained. Considering the ability of future prediction, the results of discretionary future approximation control can then be acquired. The agent modifies and improves the learning policy based on the combination of the short-sight and long-sight approximations of the future. Hence, the previous historical information combines with future information can increase the learning efficiency.
\item The comprehensive experiments demonstrate that the results of control obtained by the proposed RVL show better performances compared with the state-of-the-art control algorithms for fed-batch process.
\end{itemize}

The organization of this paper is summarized as follows. In Section II, the related work about control algorithms and RL is reviewed properly. The proposed method is introduced in Section III. Then, the details of our experiments and the discussions is presented in Section IV. Finally, the conclusion and future work are given in Section V.

\section{Related Works}
As a conventional treatment of chemical process, the fed-batch process brings in high-profile exploitation, while the product costs and desired product quantity are the major control challenges. To solve this problem, a better control policy is expected. With the development of modern technology, the control and optimization methods started to be applied in the fed-batch process in recent years. For instance, many theoretical works paid attention to step profiles to resolve the optimization issues for the fed-batch process \cite{at} \cite{to} \cite{gold}. Generally, the piecewise parameterization by the mean of linear polynomials is another kind of approach \cite{bang} \cite{tai}. The convenience of using such a smooth continuous feeding profile was marked by Martinez \textit{et al.} \cite{mar}. The feed rates were parameterized by the sinusoidal functions developed by Ochoa \cite{ocha}. The predictive control was also applied to control and optimize the fed-batch process \cite{cra} \cite{del} \cite{dew}. However, the online determination and control of processing variables are not straightforward in the initial stage. After a period of development, it is still inefficient considering that there are plenty of works to take and analyze the samples. The reversibility and uncertainty of the processing models can influence the control performances and implementations in real world.

With the development of machine learning/deep learning \cite{duan2020sofa,wang2022self,long2017zero}, there is plenty of research focusing on finding an alternative method to replace the traditional optimal control approaches. As a model-free algorithm of machine learning, RL was noticed and experienced rapid development in control area. The agent can find an optimal learning policy by a state-action value function based on the classic Q-learning \cite{caql}. To increase the efficiency of RL, Hausman \textit{et al.} \cite{hau}, Florensa \textit{et al.} \cite{flo} and  Kearns \textit{et al.} \cite{kea} explored the latent models. In addition, Gupta \textit{et al.} \cite{metar} applied the gradient-based fast adaptation algorithm to acquire exploration policy through using prior information. Garcia \textit{et al.} \cite{meta-mdp} applied the meta strategy into Markov decision process (MDP) to obtain an optimal exploration strategy. Later, several kinds of methods combined with RL were proposed to further improve the overall performances. Mnih \textit{et al.} \cite{mnih} proposed a Deep Q-network (DQN) to estimate the state-action value function. Double DQN was then estimated \cite{van} based on DQN to solve the problem of over-estimation of previous Q-network. After that, the state value and advantage value were predicted through the separated Q-network from Dueling Network explored by Wang \textit{et al.} \cite{wang}. The strength of DQN was combined with constrained optimization approach by the Optimally Tightening method by He \textit{et al.} \cite{he}. Harutyunyan \textit{et al.} \cite{haru} and Munos \textit{et al.} \cite{mu} combined on-policy samples into off-policy learning targets by $Q^{*}(\lambda)$ and Retrace$(\lambda)$. Fortunato \textit{et al.} \cite{for} proposed a Noisy-Net to increase the ability of exploration by adding noise into the parametric model during the learning progress. Distributional RL \cite{bell} learned a value function using full distribution instead of expected values. Pritzel \textit{et al.} \cite{pri} proposed a neural episodic control to generate semi-tabular representation and retrieve fast-updating values by context-based lookup for action selection. Lin \textit{et al.} \cite{lin} improved the performance of DQN and proposed an episodic memory deep Q-network by distilling information of the episodic memory. Despite the success, these methods still need to combine different algorithms with RL, and thus, DQN relied on the open environment which only considers the prior experience without future information. In addition, treating the neural networks as a state-action value function cannot leverage future information to guide the learning of RL agent.

As one of the most important algorithms in multi-agent system, multi-agent reinforcement learning (MARL) gained traction recently with various successful applications. For example, Littman \cite{litt,litt1} studied MARL in the context of Markov games. Similarly, Hu \textit{et al.} \cite{nash}, Lauer \textit{et al.} \cite{lau} and Arslan \textit{et al.} \cite{ars} applied MARL in the game learning. Jaderberg \textit{et al.} \cite{human} developed a tournament-style evaluation in 3D multiplayer games, while Bard \textit{et al.} \cite{bard} applied MARL in Hanabi as a new benchmark. Foerster \textit{et al.} \cite{foer} presented the Bayesian action decoder(BAD) as a new public belief MDP. Lee \textit{et al.} \cite{lee} proposed a policy evaluation with a linear approximation and actor-critic to improve the performance of MARL. Many efforts then concentrated on deep neural networks as a functional approximator in MARL \cite{foer1,foer2,gu,low,omi,ngu}. The relative over-generalization problem was tackled through developing a Multi-agent Soft Q-learning in continuous action spaces by Wei \textit{et al.} \cite{wei2018,wei2016}. In addition, other works like CommNet \cite{suk}, ATOC \cite{jiang2018} and SchedNet \cite{kim2019} focused on exploiting an inter-agent communication. Son \textit{et al.} \cite{qtr} proposed QTRAN to acquire a more general factorization and thus increasing the application range for MARL. Wai \textit{et al.} \cite{wai2018} applied a double averaging scheme to optimize the performance of MARL. Qu \textit{et al.} \cite{qu2019} introduced a value-propagation method based on a primal-dual decentralized optimization strategy in MARL. Liao \textit{et al.} \cite{liao2020} applied MARL in a 3D medical image segmentation problem. However, these aforementioned works focused on the cooperation of multi-agent systems, which strictly relied on an open environment. In addition, the multi-agent reinforcement learning just interacts with the internal agents of single RL algorithm, which cannot interact with agents of other algorithms.

The previous MARL and DQN have been applied successfully in various applications. However, the combination method of the proposed algorithm (namely RVL) is different from them, which involves the virtual part, basic part and cooperation part. Specifically, both virtual part and basic part can be applied with much flexibility. For example, the virtual part can exploit a traditional neural network and other models like practical swarm optimization (PSO) control method, fuzzy control approach, TD model, Sara learning, Q-learning, DQN, and MARL; imitation learning and deep recurrent Q-learning algorithms can be used in the basic part. The proposed RVL is general but very effective, which can be creatively used in a wide range of methods. To show the advantages of RVL, the virtual part and the basic part will be applied with both popular and simple prediction models and improved Q-learning method \cite{ccta}. When combined with RVL, the new model consistently outperforms the original model.

\section{Methodology}
In this paper, the proposed RVL is expected to control a fed-batch process, while the main control task is to maximize the final quality. Specifically, the number of the desirable productions can be denoted as $C_t = [c_1, c_2, ..., c_{t}]$ by a sequence of control signals $u_t = [u_1, u_2, ..., u_{t}]$. 
For RVL, the virtual space equals to the virtual environment, which can directly replace the real environment to interact with the agent of control algorithms as the basic part. Let $I_e$ be the virtual space of the virtual part, $B$ be the basic function of the basic part, and $RV$ be the final algorithm part. $I_e$, $B$ and $RV$ can be described in RVL as:
\begin{equation}
\label{eqn_example}
L^f(RV) = L^v(B_v|I_e) \circ L^r(B_r|R_e),
\end{equation}
where $L^f(RV)$denotes the optimized final learning policy, which can be acquired by a virtual learning policy $L^v(B_v|I_e)$ and a real learning policy $L^r(B_r|R_e)$; $\circ$ represents the element-wise product; $R_e$ is the real environment space. Therefore, a virtual space $I_e$ can create a virtual environment of fed-batch process in the virtual part. The basic functions $B_v$ and $B_r$ can interact with both virtual and real environments to get a virtual learning policy $L^v(B_v|I_e)$ and a real learning policy $L^r(B_r|R_e)$, and further achieve the cooperation with each other to obtain a final learning policy $L^f(RV)$. A better control signal $u$ is also given to control the fed-batch process: $L^f(RV) \rightarrow u_t = [u_1,u_2, ..., u_{t}] \rightarrow C_t = [c_1, c_2, ..., c_{t}]$.

%
\begin{figure*}[htbp]
\centering
\begin{minipage}[t]{0.8\textwidth}
\centering
\includegraphics[width=15cm, height=8.5cm]{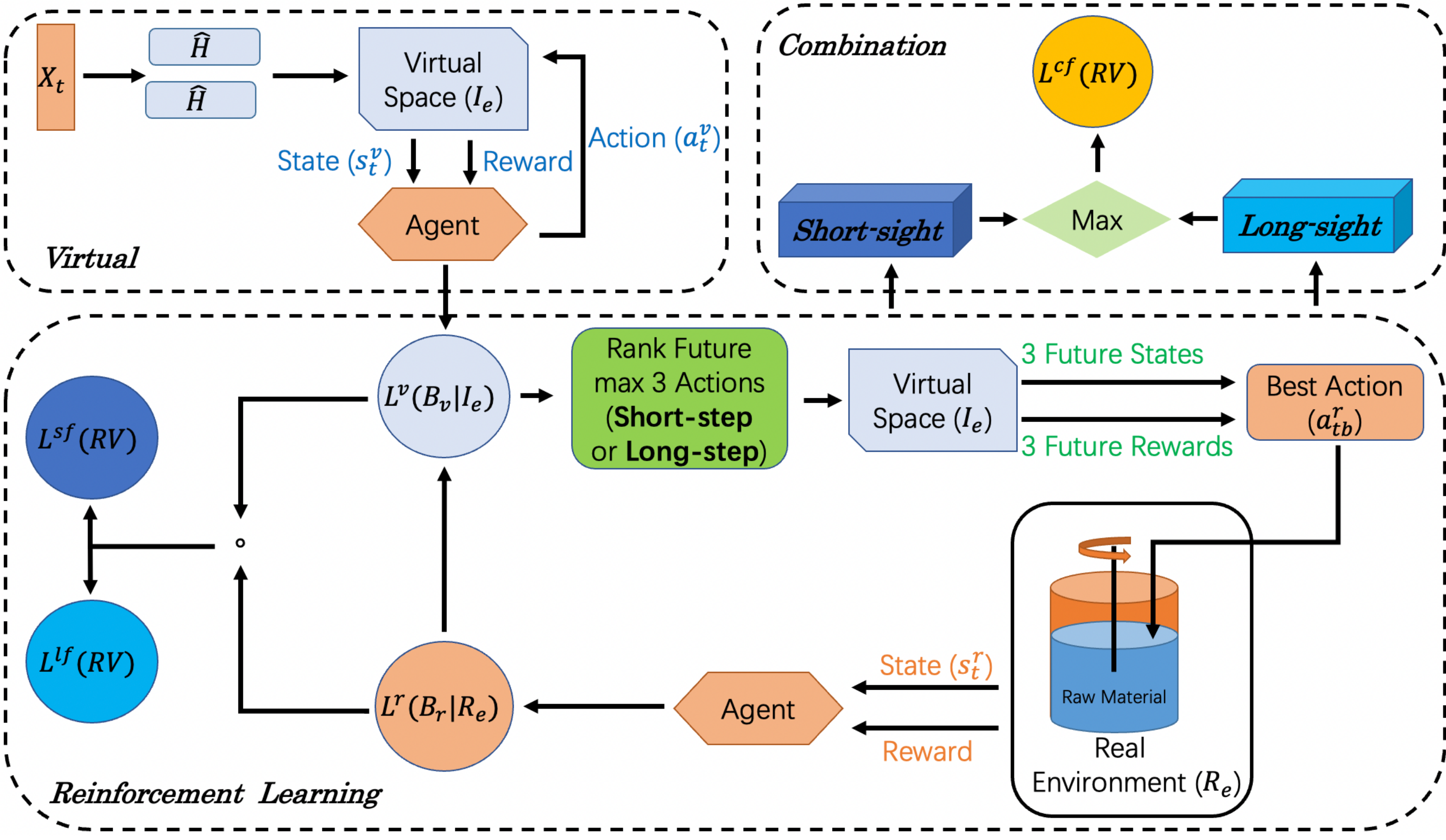}
\caption{The detailed structure of our Reinforcement Virtual Learning (RVL): The virtual learning policy can be acquired by the virtual part, which interacts with a real agent by different steps to obtain the different real learning policies. After that, they are combined to get the final learning policy.}
\end{minipage}
\end{figure*}

\subsection{Virtual Space}
An important element of RVL is the virtual space, which can create a virtual environment to interact with the agent of the basic part. With the development of the prediction models, several advanced algorithms were proposed, \textit{e.g.}, RNN, which is still the most popular prediction model so far. Plenty of improved models have then been proposed based on RNN, such as Elman Network, Jordan Network, Bi-directional Long Short-term Memory Network (BiLSTM), Gated Recurrent Unit (GRU), and Long Short-term Memory Network (LSTM) \cite{rnnoverview}. Compared with the traditional RNN, these approaches have some modifications, involving gates, memory cells, and hidden states for LSTM. Specifically, based on these developments, when LSTM resolves the time-series data, it shows a better performance compared with the traditional RNN.

For fed-batch process, the short-term reaction time affects future long-term reaction. Considering a fact that both short-term and long-term information are important, and as an advanced algorithm in RNN, LSTM is resonable to model the fed-batch processes. The gates of LSTM, as the most important component to capture valid information and store them into the memory cell, the prediction of method may benefit with higher accuracy under LSTM. 

We model a virtual space $I_e$ by an LSTM model $\hat H$ with the historical data $X_t$:
\begin{equation}
\begin{aligned}
\label{eqn_example}
I_e(S) &= \hat H | X_t(A)
= \sigma(W_{y}h_{t} + b_{y}) | X_t(A),
\end{aligned}
\end{equation}
where $\sigma$ is the sigmoid activation function, $h_t$ is the hidden state, $W_{y}$ and $b_{y}$ are the weight and bias, $A$ is the action space, and $S$ denotes the state space. The current action and the next state can be indicated by $a_t$ and $s_{t+1}$, respectively,
\begin{equation}
\label{eqn_example}
a_{t}\mathop{(}\limits a_{t} \in A) \rightarrow I_{e}(S,A) \rightarrow s_{t+1}\mathop{(}\limits s_{t+1} \in S),
\end{equation}
where the next state $s_{t+1}$ can be obtained through $I_e$ model by the selected current action $a_t$.

\subsection{Reinforcement Virtual Learning (RVL)}
\subsubsection{Virtual Leaning Policy}
This part provides the strategy of the interactions between the virtual space and RL agent. In terms of the modelled virtual space $I_e$, the agent of RL can generate the virtual state after interaction with $I_e$. Then, a virtual learning policy $L^v(B_v|I_e)$ can be acquired through a virtual basic function $B_v$:
\begin{equation}
\begin{aligned}
\label{eqn_example}
B_v = E \begin{Bmatrix}
\hat E_t|I_e (s_t^v, a_t^v)\mathop{,} \limits s_t^v \in S\mathop{,} \limits a_t^v\in A)
\end{Bmatrix}.
\end{aligned}
\end{equation}

Here, $s_t^v$ and $a_t^v$ denote the virtual state and the action, respectively. $\hat E_t$ represents the expected reward:
\begin{equation}
\begin{aligned}
\label{eqn_example}
\hat E_t = \sum_n^\infty \gamma^{n} r_{t+n},
\end{aligned}
\end{equation}
where the expected gains are denoted by $r_{t+n}$ and $\gamma$ $(0<\gamma<1)$ is the discount factor. Following Eq. (4) and Eq. (5), the virtual learning policy $L^v(B_v|I_e)$ can be described as
\begin{equation}
\begin{aligned}
\label{eqn_example}
L^{v}(B_{v}|I_{e}) \leftarrow B_{v}(s_{t}^{v},a_{t}^{m_{k}v}) +\alpha [r_{t+1}^{v} + \\ \gamma^{v}\max\limits_{a_{t}^{m_{k}v}\in A} B_{v}(s_{t+1}^{v},a_{t}^{m_{k}v}) -B_{v}(s_{t}^{v},a_{t}^{m_{k}v})],
\end{aligned}
\end{equation}
where $\alpha$ $(0<\alpha<1)$ indicates the learning rate, and $a_{t}^{m_{k}v}$ describes that a virtual action $a_t^v$ can be executed $m$ time steps in $k^{th}$ period based on SMSA \cite{ccta}. $r_{t+1}^v$ denotes the virtual expected benefits. The maximum virtual value at next virtual state $s_{t+1}^{v}$ is then represented by $\max\limits_{a_{t}^{m_{k}v}\in A} B_v(s_{t+1}^{v},a_{t}^{m_{k}v})$. Considering that the agent can interact with different environments, the weight of benefits is therefore distinguishable for RVL in different environments. Following this principle, we set different discount factors, where $\gamma^v$ represents the virtual discount factor in a virtual environment.

\subsubsection{Real Leaning Policy}
It is worth noting that the agent can acquire a virtual learning policy with a virtual environment, which means RL can be learnt in an unknown and uncertain environment. Based on this, the learned virtual learning policy can further guide the agent to learn a real learning policy $L^r(B_r|R_e)$, when the agent interacts with a real space $R_e$. 
Specifically, interacting with a real environment, the current best real action $a_{tb}^{r}$ at the current real state $s_{t}^{r}$ can be predicted based on the results of future steps by a virtual learning policy $L^v(B_v|I_e)$ combined with a virtual environment. 

For instance, three actions $a_{t1}^{v}$, $a_{t2}^{v}$, $a_{t3}^{v}$ at the current state $s_t^{v}$ can be obtained based on the virtual learning values $B_v(s_t^{v}, a_{t1}^{v}), B_v(s_t^{v}, a_{t2}^{v}), B_v(s_t^{v}, a_{t3}^{v})$ of a virtual learning policy $L^v(B_v|I_e)$ in terms of maximum to minimum:
\begin{equation}
\begin{aligned}
B_v(s_t^{v}, a_{t1}^{v}), B_v(s_t^{v}, a_{t2}^{v}), B_v(s_t^{v}, a_{t3}^{v})\leftarrow 
\\ \max\limits_{A_{t}^{v}\in A} B_v(s_{t}^{v}, A_{t}^{v})|L^v(B_v|I_e),
\end{aligned}
\end{equation}
where $A_{t}^{v}$ indicates all possible actions at state $s_{t}^{v}$. Based on the virtual learning policy, the agent can know several suitable actions in each state. In this paper, three suitable actions are enough for the task. However, the agent cannot immediately determine the best action from them. The agent needs to select three actions to interact with the virtual environment to reach three different next-states $s_{t1}^{v}, s_{t2}^{v}, s_{t3}^{v}$, respectively. After that, the agent can follow $L^v(B_v|I_e)$ to reach three different future states of $N$ steps $s_{N1}^{v}, s_{N2}^{v}, s_{N3}^{v}$. Different future-states can show the performance of control by the proposed algorithm, which can then be reflected by the expected benefit (reward) of each different state:
\begin{equation}
\begin{aligned}
\resizebox{80mm}{2.6mm}{$ r_{N}^{v}(s_{N}^{v}) = \max (r_{t}^v+(r_{N}^{v}(s_{N1}^{v}, a_{t1}^{v}), r_{N}^{v}(s_{N2}^{v}, a_{t2}^{v}), r_{N}^{v}(s_{N3}^{v}, a_{t3}^{v})) |L^v(B_v|I_e)$},
\end{aligned}
\end{equation}
where $r_{N}^{v}(s_{N}^{v})$ is the maximum reward obtained after $N$ future steps. If the maximum reward is $r_{N}^{v}(s_{N1}^{v}, a_{t1}^{v})$, the best state is $s_{N1}^{v}$, which means the best action $a_{tb}^{r}$ is $a_{t1}^{v}$ at state $s_t^{v}$ and $s_{t}^{r}$. Following this principle and the basic function $B_r$ in real space $R_e$, the virtual learning policy is similar to that of $L^r(B_r|R_e)$ to learn:
\begin{equation}
\begin{aligned}
L^r(B_r|R_e)\leftarrow B_r(s_{t}^{r},a_{tb}^{m_{k}r}) +\alpha [r_{t+1}^{r} + \\\gamma^{r}L^v(B_v|I_e)(s_{t+1}^{r},a_{tb}^{m_{k}r}) - B_r(s_{t}^{r},a_{tb}^{m_{k}r})],
\end{aligned}
\end{equation}
where the real expected benefits is denoted as $r_{t+1}^{r}$. The real states and actions are represented by $s_t^r$ and $a_t^r$.  In addition, the real discount factor is represented as $\gamma^r$. In the real space, the virtual learning policy $L^v(B_v|I_e)(s_{t+1}^{r},a_{tb}^{m_{k}r})$ guides the agent to learn a real learning policy $L^r(B_r|R_e)$, which gives both feedback and the cooperation about the previous $L^v(B_v|I_e)$ to obtain a new final virtual learning policy $L^v(B_v|I_e)$:
\begin{equation}
\begin{aligned}
\label{eqn_example}
L^v(B_v|I_e) \leftarrow B_v(s_{t}^{v},a_{t}^{m_{k}v}) +\alpha [r_{t+1}^{v} + \\ \gamma^{v}L^r(B_r|R_e)(s_{t+1}^{v},a_{t}^{m_{k}v}) - B_v(s_{t}^{v},a_{t}^{m_{k}v})].
\end{aligned}
\end{equation}

Finally, the optimized final learning policy function $L^f(RV)$ can be acquired by the cooperation between the learned virtual learning policy $L^v(B_v|I_e)$ and the real learning policy $L^r(B_r|R_e)$.

\subsubsection{Combination}
When the agent selects the best action at each state in a real environment, the future states of some steps can further influence the selection of the actions based on a virtual learning policy $L^v(B_v|I_e)$. Therefore, the future step is important for the proposed RVL, as well as the combination. For the future step, the states of the first step and the final step influence the choice of the best action at the current state. Here, the first step (named as 1-step) is referred to as the short-sight and the final step is denoted as the long-sight. 

The framework is summarized in Fig. 2. Specifically, through both the short-sight and the long-sight steps, a final short-sight learning policy $L^{sf}(RV)$ and a long-sight learning policy $L^{lf}(RV)$ can be obtained. After that, the maximum combination can be executed to acquire the final combination learning policy $L^{cf}(RV)$:
\begin{equation}
\begin{aligned}
\label{eqrv}
L^{cf}(RV)(s_{t}^{cf}, a_{t}^{cf}) = \max (L^{sf}(RV)(s_{t}^{sf}, \\
a_{t}^{sf}), L^{lf}(RV)(s_{t}^{lf}, a_{t}^{lf})).
\end{aligned}
\end{equation}

For RVL, the virtual space can interact with a basic function online to obtain the virtual knowledge. The learned information can further guide the real agents to learn real knowledge, when the agents interact with a real environment. The agents acquire useful knowledge through the virtual knowledge, thereby improving the efficiency of exploration in a real environment. The feedback of the real knowledge modifies the virtual knowledge, such that more accurate virtual knowledge can help the real agents to acquire better real knowledge. In this work, the real knowledge can be obtained effectively, resulting in the better performances for the original algorithms. Furthermore, RL can be applied directly without certain and known environments as the proposed virtual space. We summarize the proposed RVL algorithm in Algorithm 1.
\begin{algorithm}[t]
\caption{Reinforcement Virtual learning (RVL)} 
\hspace*{0.02in} {\bf Input:} 
previous data $X_t$, learning rate $\alpha$, virtual \\ discount factor $\gamma^v$, real discount factor $\gamma^r$, action space $A$, $m$ time steps $m$ of action, period $k$. $Z$ denotes the total training time of virtual space $I_e$. $O$ is whole training time of the virtual learning policy and real learning policy.\\
\hspace*{0.02in} {\bf Output:} 
Combined policy $L^{cf}(RV)$
\begin{algorithmic}[1]
\State Initial all parameters 
\For{$z = 1, ..., Z$} 
    \State Train $\hat H$ with $X_t$ with $\hat H$.
  		\State Get virtual space $I_e$.
 \EndFor \State\noindent\textbf{end for}
 \State Training $L^{sf}(RV)$:
 \For{$o = 1, ..., O$}
 		\If{$mod(J,p) \sim = 0$}
        		\For{agent $t = 1, ... , T$}
                		\State Randomly obtain $m$ and $k$.
                        \State Apply action $a_{t}^{m_{k}v}$ under model $I_e$.
                        \State $s_t^v \rightarrow s_{t+1}^v$.
                        \If{$J < p$}
                        		\State Update $L^v(B_v|I_e)$ using Eq. (6).
                        \Else
                        		\State Update $L^v(B_v|I_e)$ using Eq. (10).
                        \EndIf \State\noindent\textbf{end if}
                 \EndFor\State\noindent\textbf{end for}
        \Else
        		\For{agent $t = 1, ..., T$}
                		\State Randomly obtain $m$ and $k$.
                        \State $r_{N}^{v}$(short), best $a_{tb}^{m_{k}r}$ using Eq. (7) and Eq. (8)
                        \State Apply $a_{tb}^{m_{k}r}$ under real environment $R_e$.
                        \State $s_t^r \rightarrow s_{t+1}^r$.
                        \State Update $L^r(B_r|R_e)$ using Eq. (9).
                \EndFor\State\noindent\textbf{end for}
        \EndIf\State\noindent\textbf{end if}
 	\EndFor\State\noindent\textbf{end for}
 	\State\Return $L^{sf}(RV)$.
 	\State Training $L^{lf}(RV)$:
    \State Repeat steps 7 to 28, replace $r_{N}^{v}$(short) with $r_{N}^{v}$(long).
    \State\Return $L^{lf}(RV)$.
    \State Update $L^{cf}(RV)$ using Eq. (11).
    \State\Return  $L^{cf}(RV)$.

\end{algorithmic}
\end{algorithm}

\section{Experiments}
We design different sets of experiments to verify the performance of the proposed method. The advantages of the new algorithm can be shown directly by our control results, where the key results are analyzed below.

\subsection{Set-up for the Dataset}
\subsubsection{Fed-batch Process Model}
Although a number of control algorithms were applied in chemical processes, machine learning-based methods were explored barely in recent years. It is worth noting that machine learning-based control results are often superior to others, which means machine learning-based technique can be applied successfully in chemical processes. As a traditional process of chemical processes, the batch process is important. The main strategy is that the proposed algorithm can control it optimally as shown in our experiments.

The fed-batch process is a classical batch process, and therefore, we apply it in this work. This fed-batch process is described as follow:
\begin{equation}
\label{eqn_example}
A+B\stackrel{k_{1}}{\longrightarrow}C,
\end{equation}
\begin{equation}
\label{eqn_example}
B+B\stackrel{k_{2}}{\longrightarrow}D,
\end{equation}
where the reactants $A$ and $B$ are the raw materials; $C$ and $D$ are the desirable productions and the undesirable by-products, respectively. The reactant $B$ can be added into the reactor gradually, to prevent the fast formation of the undesirable by-products $D$ during the specified batch time $t_{f}=120$ $min$.

In the fed-batch process, the main control purpose is that the desirable products $C$ should be acquired as much as possible, while the undesirable products $D$ should be kept at the lowest quantity in the whole reaction batch time, where the total volumes $V$ cannot exceed 1 $m^3$.

In the control task, the concentration of reactant $B$ is added in a feed stream with concentration $b_{feed}=0.2$. The following fed-batch process model is developed based on the material balances and the reaction kinetics:
\begin{equation}
\begin{aligned}
\label{eqn_example}
&\frac{d[A]}{dt} = -k_{1}[A][B]-\frac{[A]}{V}u, \\
&\frac{d[B]}{dt}=-k_{1}[A][B]-2k_{2}[B]^{2}+\frac{b_{feed}-[B]}{V}u, \\
&\frac{d[C]}{dt}=-k_{1}[A][B]-\frac{[C]}{V}u, \\
&\frac{d[D]}{dt}=2k_{2}[B]^{2}-\frac{[D]}{V}u, \\
&\frac{d[V]}{dt}=u.
\end{aligned}
\end{equation}

The concentrations of $A$, $B$, $C$ and $D$ are represented by [$A$], [$B$], [$C$] and [$D$], respectively. The volume of the materials in the reactor and the reactant feed rate are denoted by $V$ and $u$, respectively. The reaction rates are represented as $k_{1}$ and $k_{2}$, and are set to 0.5, as shown in Table I. The initial [$A$] is 0.2 moles/litter and [$V$] is 0.5. Based on the above model, a simulation program of the fed-batch process can be developed using Matlab, and the simulation is used to test the various control algorithms. In this paper, the simulation of fed-batch process is called the real reaction process.  
\begin{figure}[ht]
\centering
     \hspace{-13.6ex}
         \includegraphics[width=0.48\textwidth]{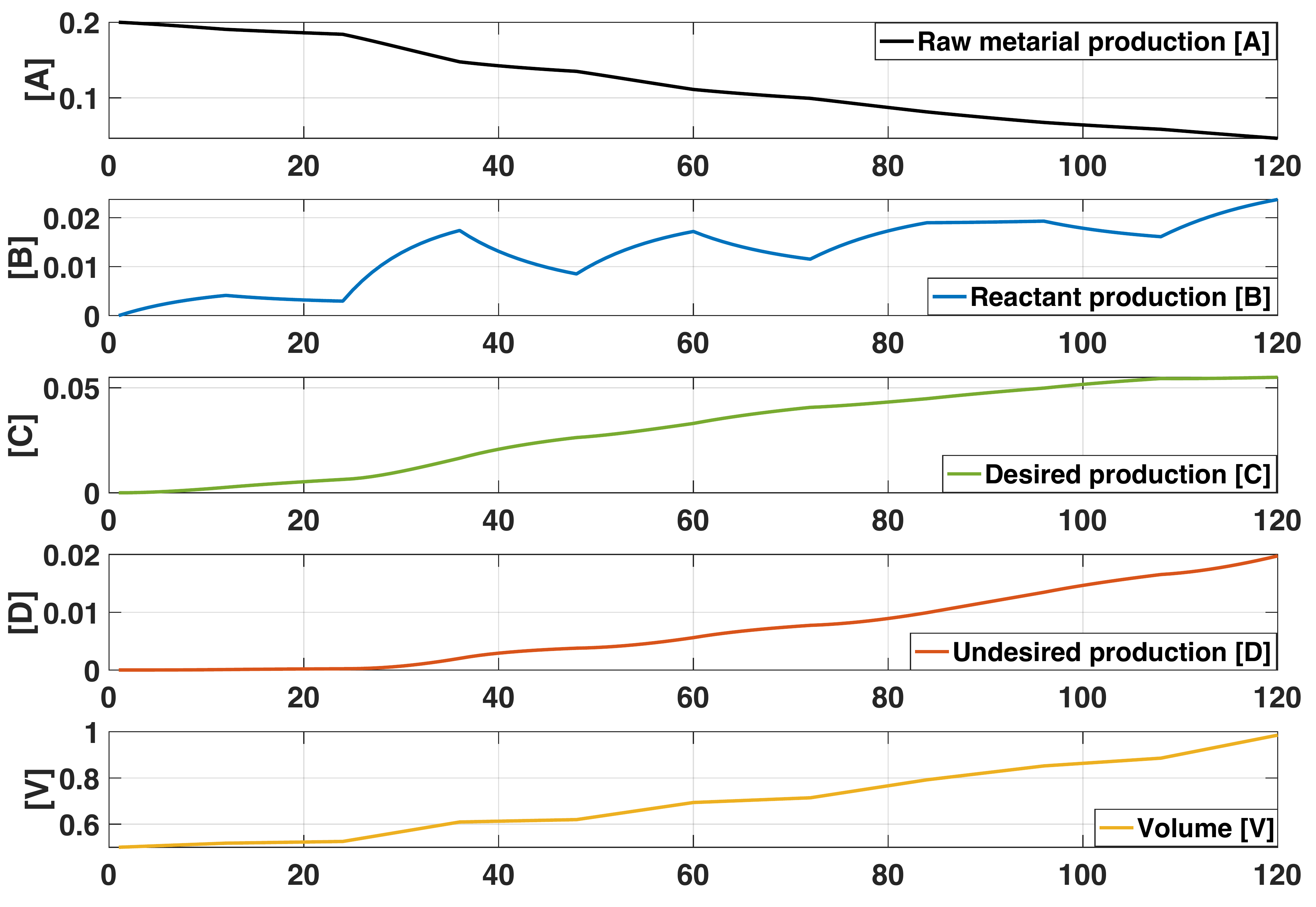}
    \hspace{-12.6ex}
    \caption{An example sequence of [A], [B], [C], [D], [V] during a reaction process based on real environment}
\end{figure}

In terms of the real reaction process and an example sequence [$A$], [$B$], [$C$], [$D$], [$V$] is shown in Fig. 3.

\subsubsection{Dataset}
The dataset is constructed by 20,000 sequences, relying on the base fed-batch process model in our experiments.

Let sequences control signals be $U_{t}^{i} = [u_{t}^{1}, u_{t}^{2}, u_{t}^{3}, ..., u_{t}^{i}]$, the desired productions be $C_{t}^{i} = [c_{t}^{1}, c_{t}^{2}, c_{t}^{3}, ..., c_{t}^{i}]$, the undesired productions be $D_{t}^{i} = [d_{t}^{1}, d_{t}^{2}, d_{t}^{3}, ..., d_{t}^{i}]$, and the constructed historical information $X_{t}$. We randomly select 15,000 sequences data as our training data, and the remaining 5,000 sequences are taken the test data.

For desired product $C$, the prediction model $\hat H$ has 100 hidden neurons in hidden state layer and the mini-batch size is set to be 20, then this model is trained by 3,000 epochs. Compared with desired product $C$ model, the prediction model of undesired product $D$ has 200 hidden neurons and the training time is 6,000 epochs.

\subsection{Reinforcement Virtual Learning Design for Fed-batch Process}
In this paper, RVL is based on the traditional RL, such that the important construction elements of RVL are similar to the traditional Q-learning. Therefore, the models of agent, state, action and reward function are vital as well.

\subsubsection{The Agent Design}
As an element of RVL, several important parameter of the RL model (e.g., learning rate $\alpha$ and discount factor $\gamma$) should be set first. For the proposed algorithm, two different learning policies will continuously interact during the learning time with two different discount factors $\gamma^v$ and $\gamma^r$. Specifically, the virtual learning policy $L^v(B_v|I_e)$ of the virtual space $I_e$ is trained by the prediction model $\hat H$, and its discount factor $\gamma^v$ influences less than that of in the real learning policy $L^r(B_r|R_e)$ of the real environment $R_e$ after $L^v(B_v|I_e)$. The discount factor $\gamma^r$ is expected to significantly affect the final learning policy $L^f(RV)$. In addition, as an essential part of RVL, the $\epsilon$-$greedy$ policy needs to be set with a suitable $\epsilon$ value. Table I denotes these parameters for our experiments.
\begin{table}[h]
\caption{Parameters used in the simulations.}
\begin{center}
\begin{tabular}{ccc}
\hline
\hline
\textit{\textbf{Variable}}                 & \textit{\textbf{Meaning}}                                                  &  \textit{\textbf{Setting}} \\ \hline 
$k_{1}, k_{2}$                  & \begin{tabular}[c]{@{}c@{}}reaction rate\end{tabular}   & 0.5     \\ 
$\alpha$                 & \begin{tabular}[c]{@{}c@{}}learning rate\end{tabular}   & 0.1     \\ 
$\gamma^v$                 & \begin{tabular}[c]{@{}c@{}}virtual discount factor\end{tabular} & 0.7    \\ 
$\gamma^r$                 & \begin{tabular}[c]{@{}c@{}} real discount factor\end{tabular} & 0.98    \\ 
$\varepsilon$   & greedy-probability                                        & 0.7     \\ \hline\hline
\end{tabular}
\end{center}
\label{tabl1}
\end{table}

\subsubsection{The State Design}
In our experiments, the main control purpose is that the desirable products [$C$] are produced as much as possible, while the undesirable by-products [$D$] should be kept at a low quantity at the end \cite{ccta}. We design the state based on this principle. During the given reaction time, each product goes through some fluctuations following the implementation of the control policy. Once the increasing rate of the desirable product concentration is high, more desired products are expected to be produced, while the increasing rate of the undesirable products should be kept low at the meantime. Following this principle, the slope of [$C$] curve should be steeper than that of [$D$] curve, to achieve a desired result in the whole reaction time. Therefore, the state can be represented by the differences in derivatives between [$C$] and [$D$] as described in Table II, where $\Delta$[$C$] and $\Delta$[$D$] are the slope of [$C$] and [$D$], respectively.
\begin{table}[h]
\renewcommand\arraystretch{2}
\caption{States of the fed-batch process}
\begin{center}
\begin{tabular}{l|l}
\hline
\hline
\textit{\textbf{Condition}}               & \textit{\textbf{State}} \\ \hline \hline
$\Delta [C]-\Delta [D]\geq 0.0008$        & $S_{1}$                 \\ 
$0.0007 \le \Delta [C]-\Delta [D]< 0.0008$ & $S_{2}$                 \\ 
$0.0006 \le \Delta [C]-\Delta [D]< 0.0007$ & $S_{3}$                 \\ 
$0.0005 \le \Delta [C]-\Delta [D]< 0.0006$ & $S_{4}$                 \\ 
$0.0004 \le \Delta [C]-\Delta [D]< 0.0005$ & $S_{5}$                 \\ 
$0.0003 \le \Delta [C]-\Delta [D]< 0.0004$ & $S_{6}$                 \\ 
$0.0002 \le \Delta [C]-\Delta [D]< 0.0003$ & $S_{7}$                 \\ 
$0.0001 \le \Delta [C]-\Delta [D]< 0.0002$ & $S_{8}$                 \\ 
$0 \le \Delta [C]-\Delta [D]< 0.0001$      & $S_{9}$                 \\ 
$\Delta [C]-\Delta [D]< 0$                 & $S_{10}$                \\ \hline\hline
\end{tabular}
\end{center}
\end{table}

\subsubsection{The Action Design}
In our experiments, [$A$] is given at the beginning of the reaction. With the adding of $u$, [$B$], [$C$], [$D$] and [$V$] are changed. Therefore, the feeding rate $u$ decides the the control signal and the action space in the range of from 0.001 to 0.009.


\subsubsection{The Expected Benefit Function Design}
For any algorithms of RL, the design of the benefit is one of the most important part. Actually, the benefit and the punishment of the expected benefit function can directly influence the learning performance of the algorithm, resulting in a flexible design of the expected benefit function. 

The agent can predict future results accurately by the virtual part, leading to the improvement of the the accuracy of the expected benefit and its selection of action in the whole learning process for the proposed RVL. In this case, a direct and simple benefit function is approximated, with the design of the expected benefit function represented by a constant value based on the traditional methods. In this fed-batch process, the distribution of the expected benefit function is followed by a state space.

\subsection{The Control Results}
\subsubsection{Experimental Details}
Considering a fact that RVL creates the virtual learning policy $L^v(B_v|I_e)$, which can predict the estimated future results to indicate and interact with the real agent to further acquire a better real learning policy $L^r(B_r|R_e)$. Based on this principle, RVL will predict some future steps during the control and the learning processes. We set several experiments based on the virtual 1-step, 30-step, 50-step, 80-step and 120-step (final step). After that, the proposed combination-step experiments will be applied as well.

\subsubsection{Comparison with Other Algorithms}
To show the control performances, the control results of RVL are directly compared with other state-of-the-art control algorithms, such as the recurrent neuro-fuzzy network, traditional Q-learning, stochastic multi-step action Q-learning (SMSA) \cite{ccta}, nominal control, and minimal risk control algorithm \cite{jie}. Table III shows the results of different control algorithms.
\begin{table*}[h]
\renewcommand\arraystretch{2}
\begin{center}
\caption{The control results of RVL compared with other control algorithms}
\begin{tabular}{l|cccc|cc}
\hline
\hline
\textit{\textbf{Algorithm}}            & \textit{\textbf{{[}C{]}}} & \textit{\textbf{{[}D{]}}} & \textit{\textbf{{[}V{]}}} & \textit{\textbf{{[}C{]}-{[}D{]}}} & \textit{\textbf{({[}C{]}-{[}D{]})*{[}V{]}}} \\ \hline \hline
\textbf{Recurrent neuro-fuzzy network} \cite{jie} & 0.0559                    & 0.0304                    & 0.9900                    & 0.0355                            & 0.0351                                      \\ \hline
\textbf{Nominal control} \cite{jie}              & 0.0615                    & 0.0345                    & 0.9918                    & 0.0267                            & 0.0264                                      \\ \hline
\textbf{Minal risk} \cite{jie}                   & 0.0612                    & 0.0236                    & 1.000                     & 0.0376                            & 0.0376                                      \\ \hline
\textbf{Q-learning} \cite{ccta}                   & 0.0590                    & 0.0193                    & 0.9220                    & 0.0366                            & 0.0366                                      \\ \hline
\textbf{SMSA} \cite{ccta}                          & \textit{\textbf{0.0618}}                    & 0.0236                    & 0.9800                    & 0.0361                            & 0.0361                                      \\ \hline\hline
\textit{\textbf{RVL}}                 & 0.0614  & \textit{\textbf{0.0199}}  & \textit{\textbf{0.9254}}  & \textit{\textbf{0.0415}}          & \textit{\textbf{0.0384}}                    \\ \hline\hline
\end{tabular}
\end{center}
\end{table*}

In Table III, although more desirable productions [$C$] are produced by the nominal control and the SMSA algorithm when compared with other algorithms, more undesirable productions [$D$] are generated as well. For RVL, we note that the difference between [$C$] and [$D$] is maximum, and the difference between desired final species [$C$][$V$] and undesired final species [$D$][$V$] is also maximum. This indicates that when the proposed RVL algorithm achieves the best compared with other control algorithms.
\begin{figure}[h]
\centering
\includegraphics[width=0.24\textwidth]{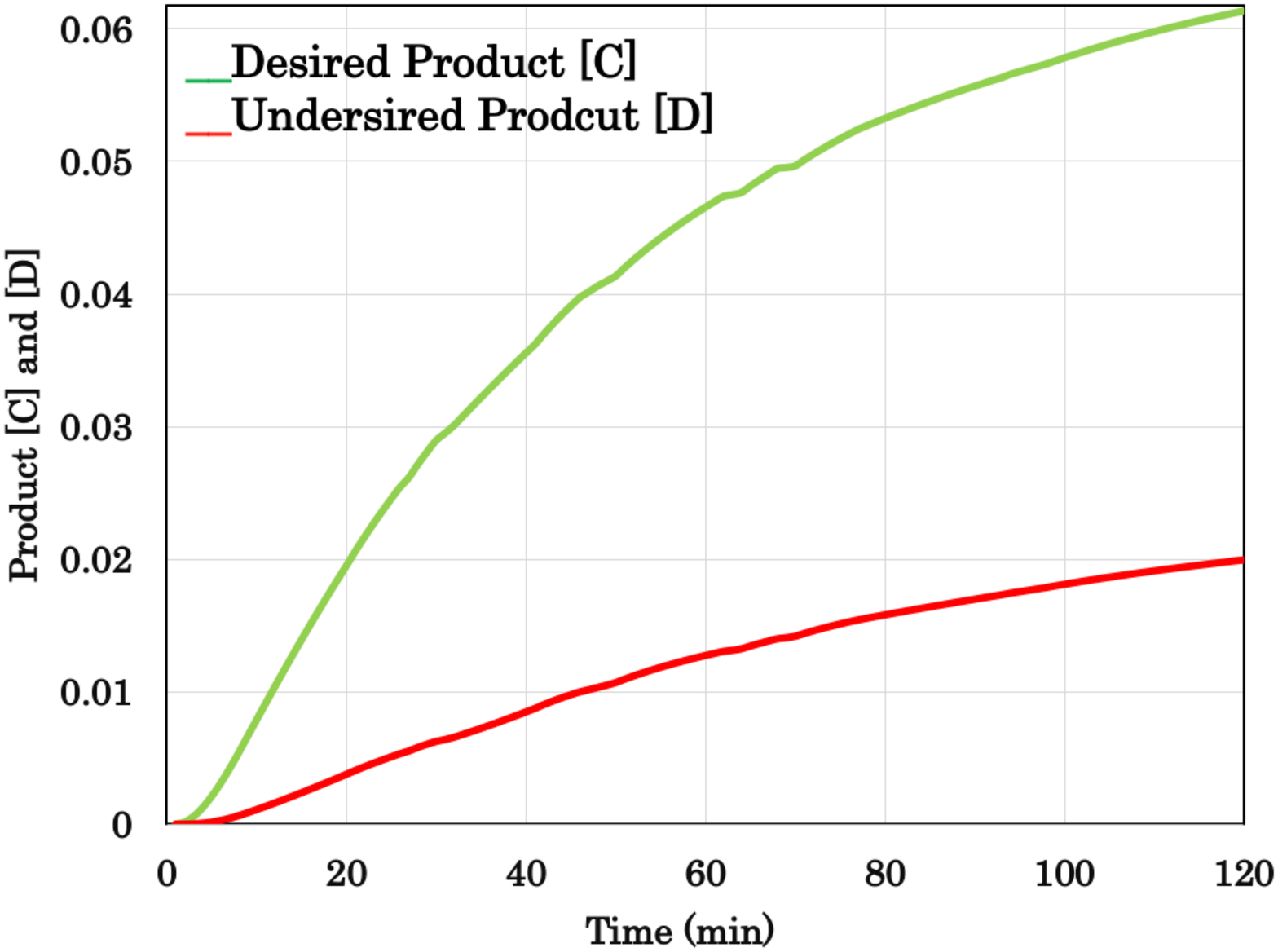}
\includegraphics[width=0.24\textwidth]{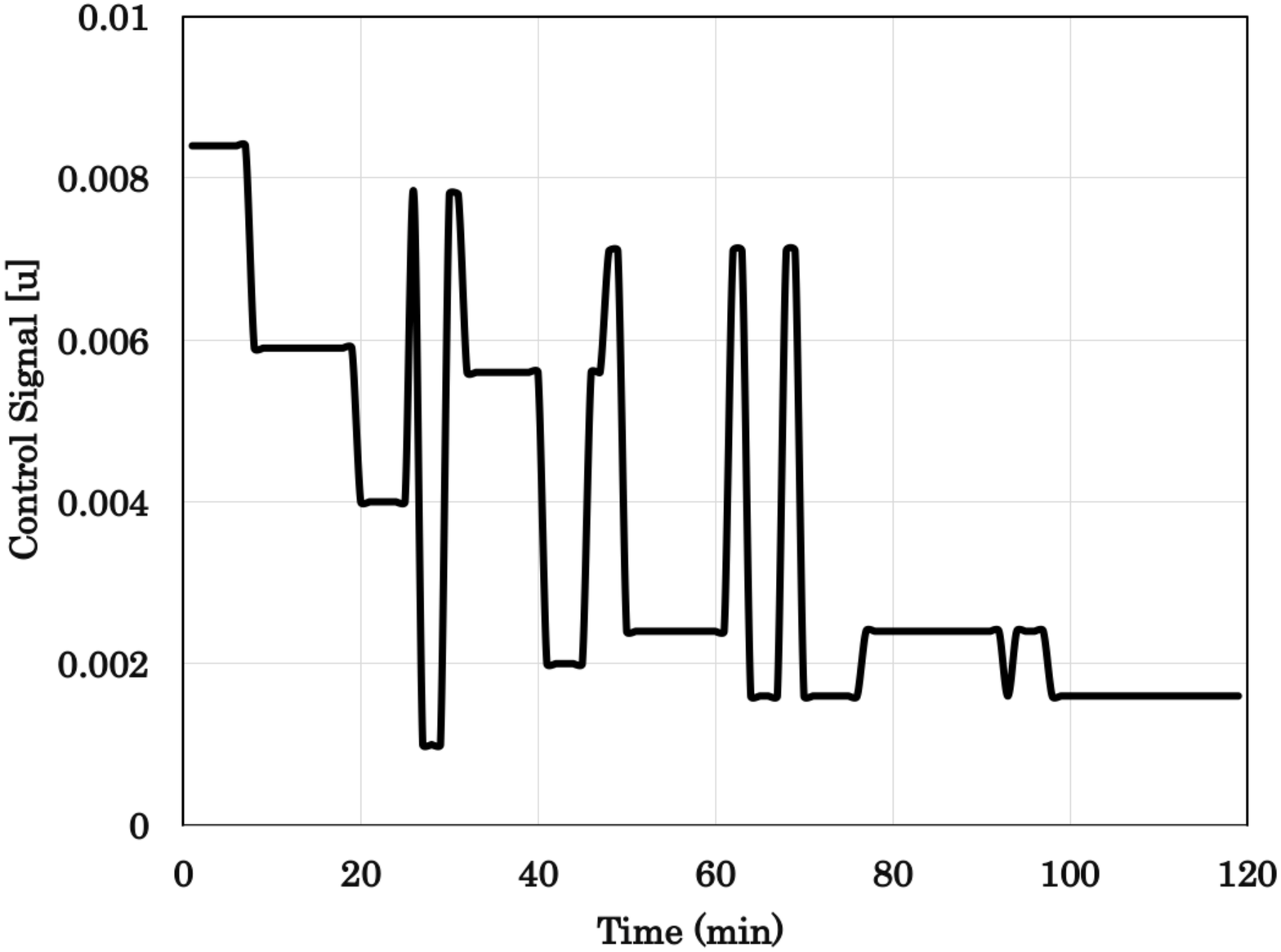}
\caption{The variation during a reaction process of the desirable products [C] and the undesirable products [D] based on RVL control (left). The control signal [u] under RVL control (right).}
\end{figure}

In summary, the final control algorithm will follow the combination of the short-long step based on RVL. The control results of [$C$] and [$D$] are described in Fig. 4, and the final suitable control signal under the proposed algorithm is shown in Fig. 4.

\subsection{Detailed Evaluations} 
\subsubsection{The Virtual Prediction Results}
%
\begin{figure}[h]
\centering
\includegraphics[width=0.24\textwidth]{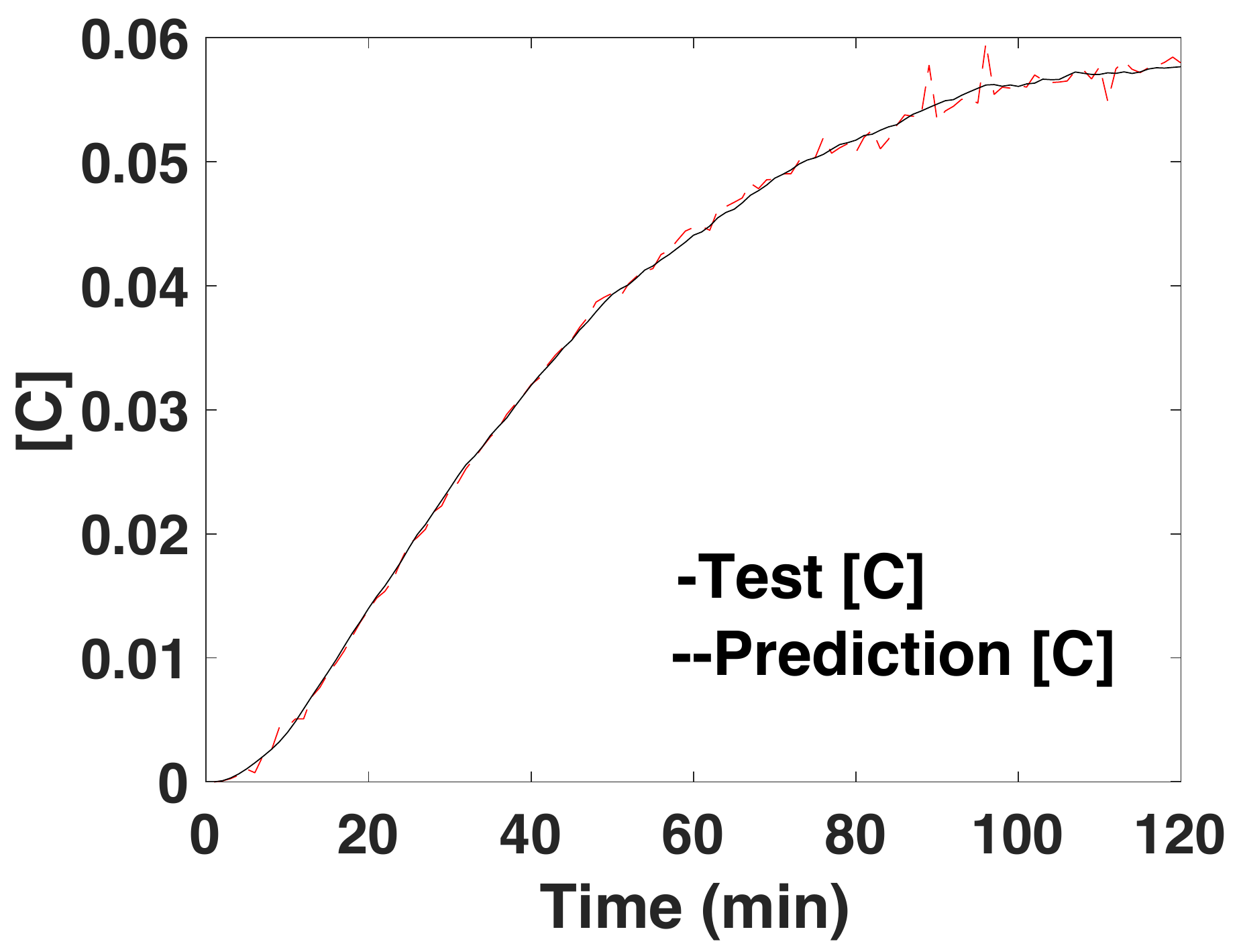}
\includegraphics[width=0.24\textwidth]{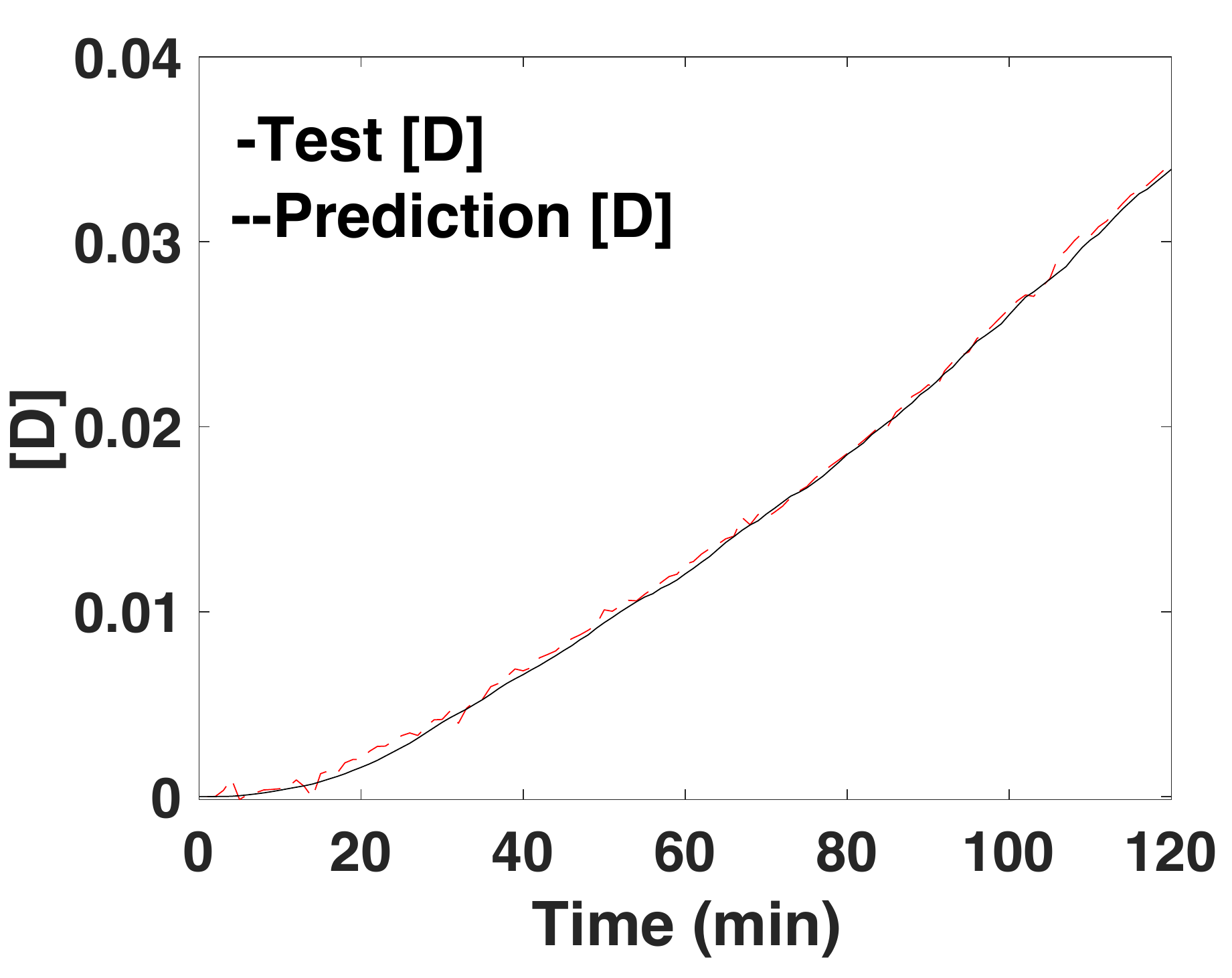}
\caption{The prediction and the ground truth of the desirable products [C] (left) and the undesirable products [D] (right).}
\end{figure}

Fig. 5 presents the prediction and the ground truth of the desirable products [$C$] and the undesirable products [$D$]. We observe that both products can be predicted accurately under the model $\hat H$.

The Root Mean Squared Error (RMSE) between the predictions and the test data of two different productions are shown in Fig. 6. It shows that the trained desirable product model $C$ and the undesirable product model $D$ under $\hat H$ can predict the real reaction process of [$C$] and [$D$] accurately.
\begin{figure}[h]
\centering
\includegraphics[width=0.24\textwidth]{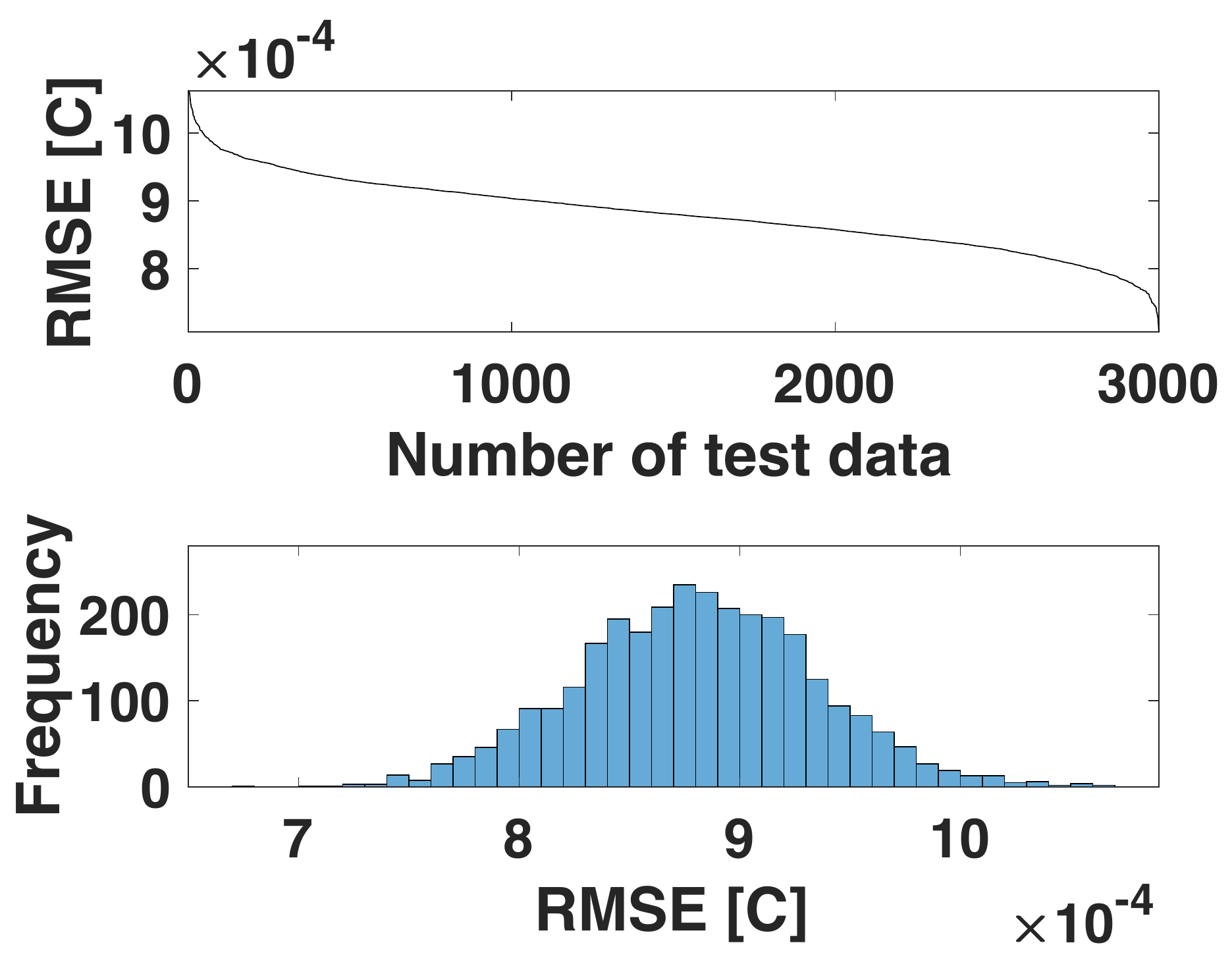}
\includegraphics[width=0.24\textwidth]{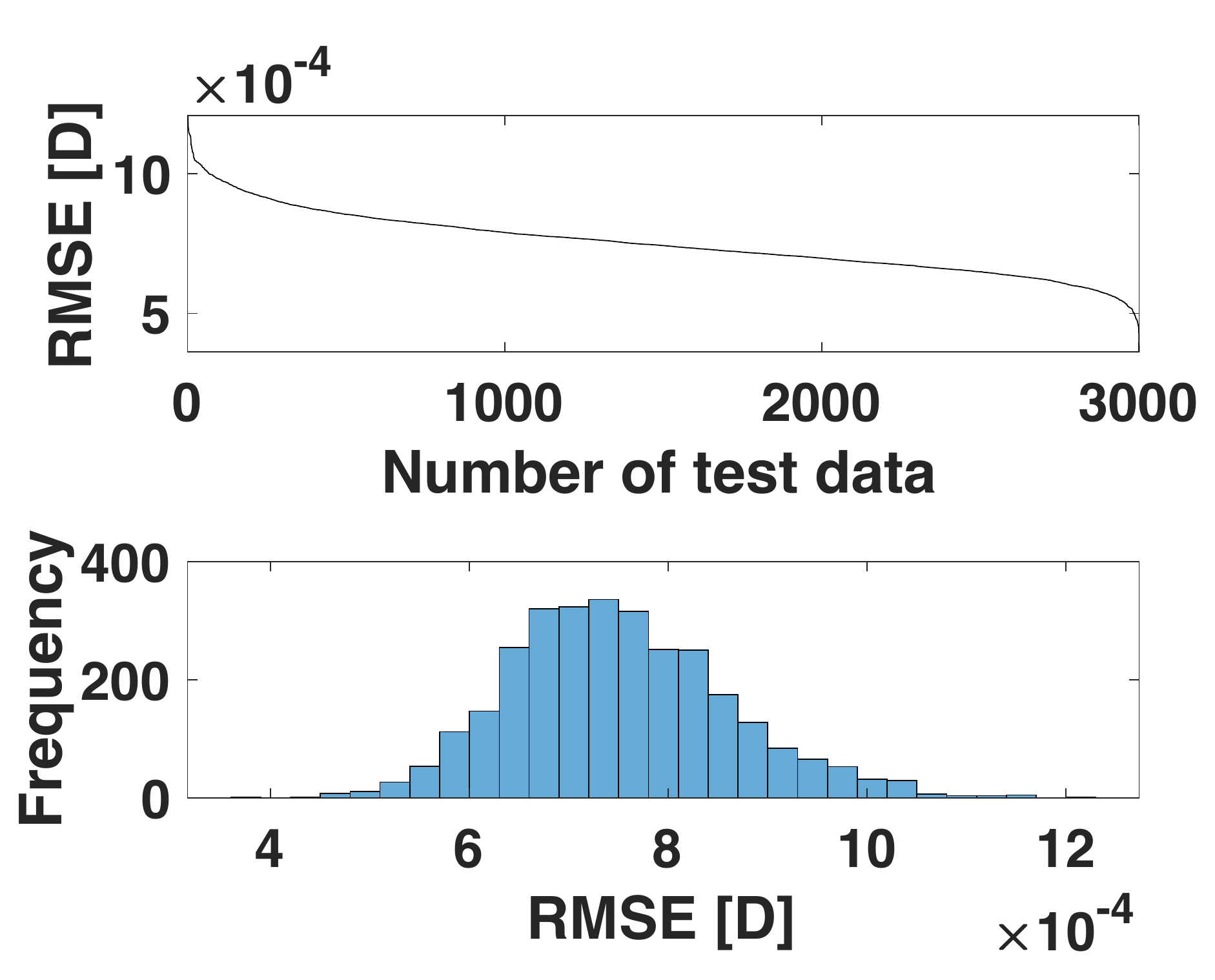}
\caption{RMSE between the predictions and the ground truth of the desirable products [C] (left) and the undesirable products [D] (right).}
\end{figure}

In summary, the model $\hat H$ shows the outstanding performance, which can clearly predict the variations of the desirable products [$C$] and the undesirable products [$D$] under different control signals for the fed-batch process. Therefore, the trained models for $C$ and $D$ can be referred as the virtual reaction process, which can replace the real reaction process, especially for learning the virtual learning policy.

\subsubsection{Impact of Step Size}
In order to describe the control performance of the proposed RVL algorithm, the results of different pure steps are shown: short-sight (1-step), immediate-sight (30-step, 50-step and 80-step) and long-sight (120-step), respectively. The control results of the combination steps of different sights are reported as follows. 

\begin{table}[]
\renewcommand\arraystretch{2}
\caption{The control results of [C] and [D] based on different pure steps.}
\begin{center}
\begin{tabular}{c|cccc}
\hline
\hline
\textit{\textbf{Algorithm}} & \textit{\textbf{{[}C{]}}} & \textit{\textbf{{[}D{]}}} & \textit{\textbf{{[}V{]}}} & \textit{\textbf{({[}C{]}-{[}D{]})*{[}V{]}}} \\ \hline \hline
\textit{\textbf{1-step}}    & 0.0606                    & 0.0182                    & 0.8999                    & \textit{\textbf{0.0381}}                    \\ 
\textit{\textbf{30-step}}   & 0.0558                    & 0.0173                    & 0.9433                    & 0.0363                                      \\ 
\textit{\textbf{50-step}}   & 0.0566                    & 0.0179                    & 0.9638                    & 0.0372                                      \\ 
\textit{\textbf{80-step}}   & 0.0579                    & 0.0218                    & 1.0000                    & 0.0361                                      \\ 
\textit{\textbf{120-step}}  & 0.0613                    & 0.0211                    & 0.9254                    & \textit{\textbf{0.0372}}                    \\ \hline \hline
\end{tabular}
\end{center}
\end{table}

Table IV indicates the results of the desirable products [$C$] and the undesirable products [$D$] based on different pure steps. Fig. 7 describes the variation curves.


\begin{figure}[h]
\centering
\includegraphics[width=0.24\textwidth]{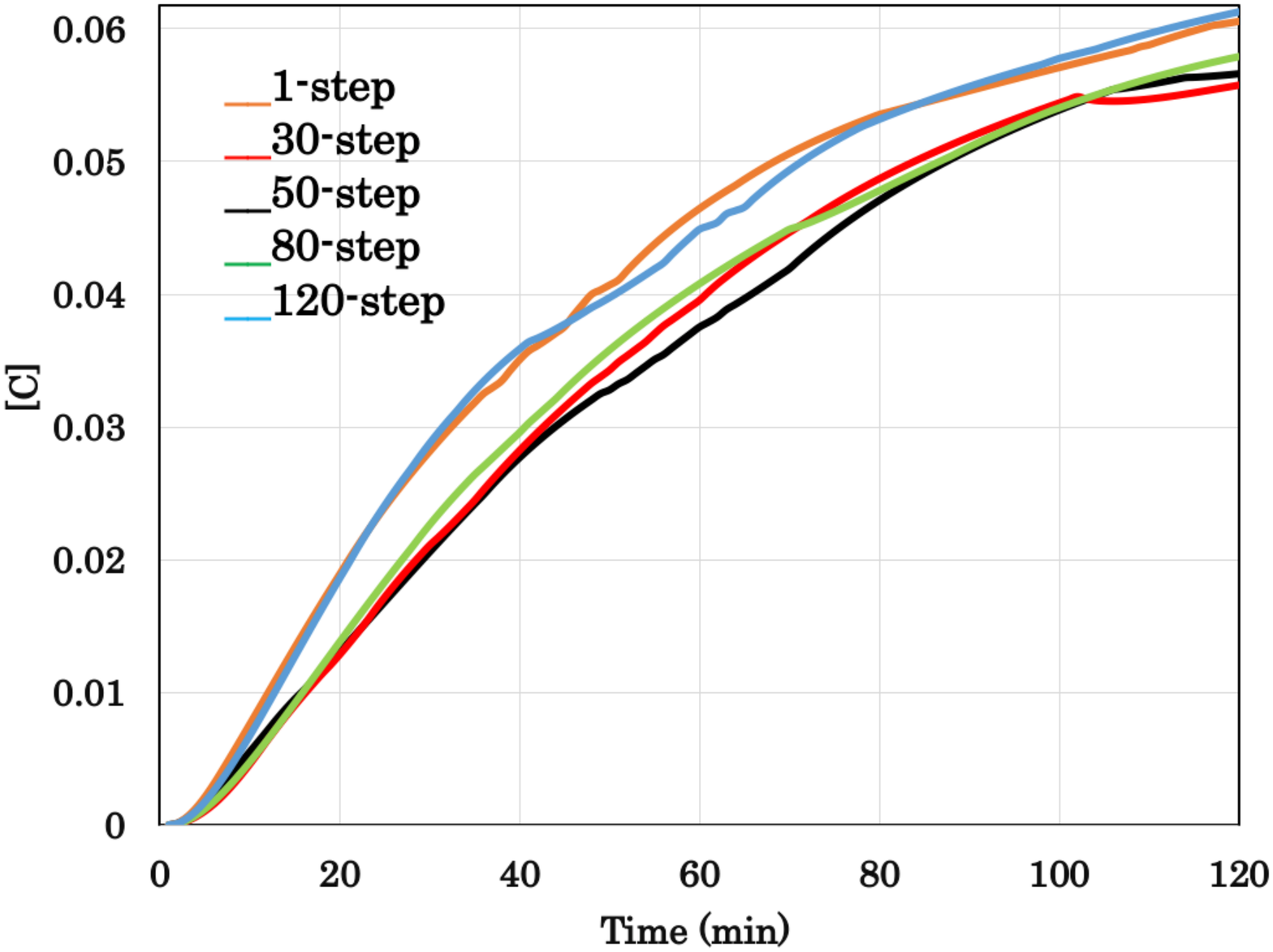}
\includegraphics[width=0.24\textwidth]{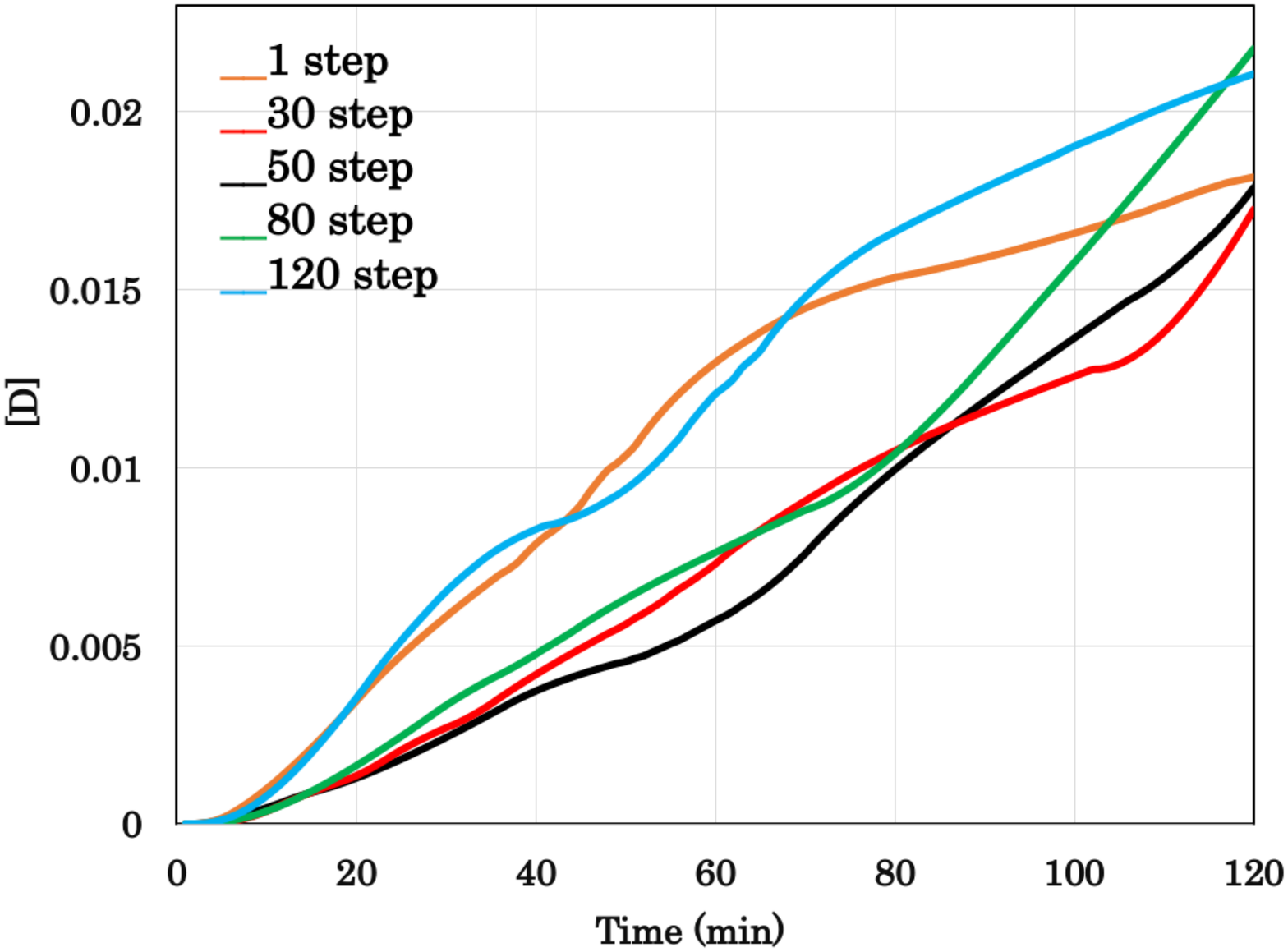}
\caption{The variation curves of [C] under different pure steps control (left) and [D] (right).}
\end{figure}

\begin{figure}[ht]
\centering
     \hspace{-13.6ex}
         \includegraphics[width=0.51\textwidth]{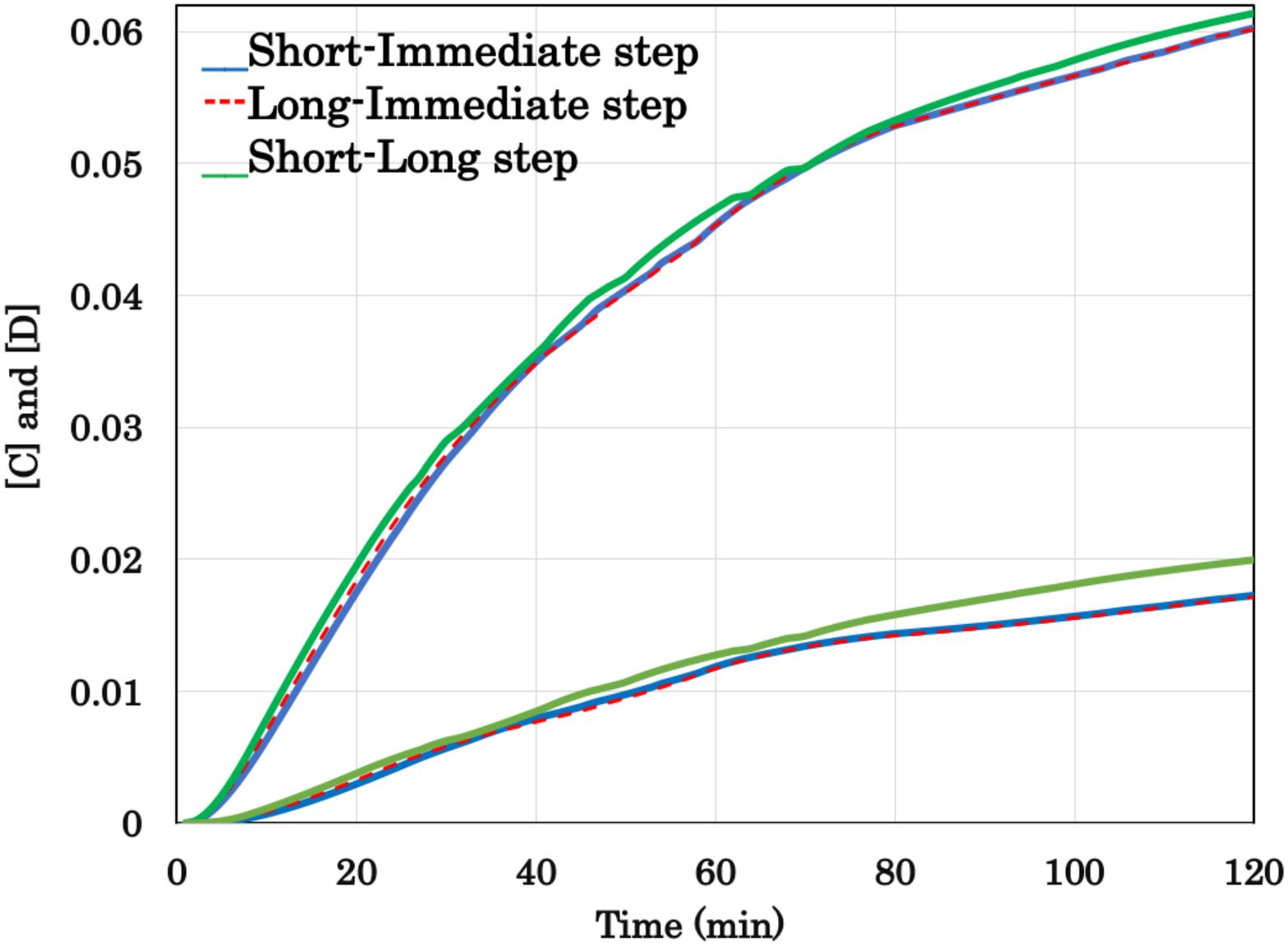}
    \hspace{-12.6ex}
    \caption{The variation curves of the desirable [C] and undesirable [D] under different combination steps of control.}
\end{figure}


\begin{table}[h]
\renewcommand\arraystretch{3}
\begin{center}
\caption{The control results based on different combination-steps}
\begin{tabular}{c|cccc}
\hline
\hline
\textit{\textbf{Algorithm}}            & \textit{\textbf{{[}C{]}}} & \textit{\textbf{{[}D{]}}} & \textit{\textbf{{[}V{]}}} & \textit{\textbf{({[}C{]}-{[}D{]})*{[}V{]}}} \\ \hline \hline
\textit{\textbf{Short-Immediate step}} & 0.0603                    & 0.0173                    & 0.8898                    & 0.0382                                      \\ 
\textit{\textbf{Immediate-Long step}}  & 0.0601                    & 0.0171                    & 0.8913                    & 0.0383                                      \\ \hline\hline
\textit{\textbf{Short-Long step}}      & \textit{\textbf{0.0614}}  & 0.0199                    & 0.9254                    & \textit{\textbf{0.0384}}                    \\ \hline \hline
\end{tabular}
\end{center}
\end{table}

In Fig. 8 and Table IV, we can observe that when the agent predicts 1-step and 120-step, more desirable productions [$C$] can be acquired compared with other steps of control, which means the short-sight and long-sight have a better control performance.

Secondly, the combination-step will be applied. In our experiments, the learning policy of short-sight will be combined with that of the immediate-short and the long-sight, respectively. When the algorithm applies the combination-step, we acquire a better performance as shown in Fig. 7 and Table V.

Specifically, there are more desirable productions [$C$] and less undesirable productions [$D$] after being applied the combination-step compared with the pure immediate-step (30-step, 50-step, 80-step) and the long-sight step (120-step).


Once the combination-step is applied, the improvement for control can be proved by the total expected benefits. When the combination-step is applied, the total reward will be increased compared with different pure steps. We demonstrate the details in Table VI.

Following Table VI, 1-step (short-sight) and 120-step (long-sight) can collect more expected benefits than immediate-sight for the pure step, which indicates that the control results of short-sight and long-sight are better as shown in Fig. 8 and Table IV. In addition, it also proves that the expected benefits can reflect the performance of RL and control results. Obviously, the combination-step can acquire more expected benefits in total compared with different pure steps, and thus, the performance of learning policy and control of combination-step will be better. Especially, the combination-steps of short-immediate and immediate-long-sight can be improved significantly compared with immediate-sight (30-step, 50-step and 80-step).  
\begin{table}[h]
\renewcommand\arraystretch{2}
\begin{center}
\caption{The total expected benefits of different steps of algorithms.}
\begin{tabular}{c|c}
\hline\hline
\textit{\textbf{Algorithm}}                   & \textit{\textbf{Total Expected Benefits}} \\ \hline\hline
\textit{\textbf{1-step (Short-sight)}}        & 33100                        \\ 
\textit{\textbf{30-step (Immediate-sight)}}    & 6500                         \\ 
\textit{\textbf{50-step (Immediate-sight)}}    & 16200                        \\ 
\textit{\textbf{80-step (Immediate-sight)}}    & 7900                         \\ 
\textit{\textbf{120-step (Long-sight)}}       & 26700                        \\ \hline\hline
\textit{\textbf{Short-Immediate Combination}} & \textit{\textbf{34400}}      \\ 
\textit{\textbf{Immediate-Long Combination}}  & \textit{\textbf{33400}}      \\ 
\textit{\textbf{Short-Long Combination}}      & \textit{\textbf{30800}}      \\ \hline\hline
\end{tabular}
\end{center}
\end{table}

Although the total expected benefits of 1-step, short-immediate, and immediate-long combination-step are greater than the short-long combination-step, the final control result of short-long combination-step is the best. The reason is that the expected benefits can be acquired easily in the previous and the immediate reaction time (short-sight and immediate-sight) compared with the latter reaction time (long-sight) in terms of the state. The expected benefit function is shown in Fig. 9.
\begin{figure}[ht]
\centering
     \hspace{-13.6ex}
         \includegraphics[width=0.51\textwidth]{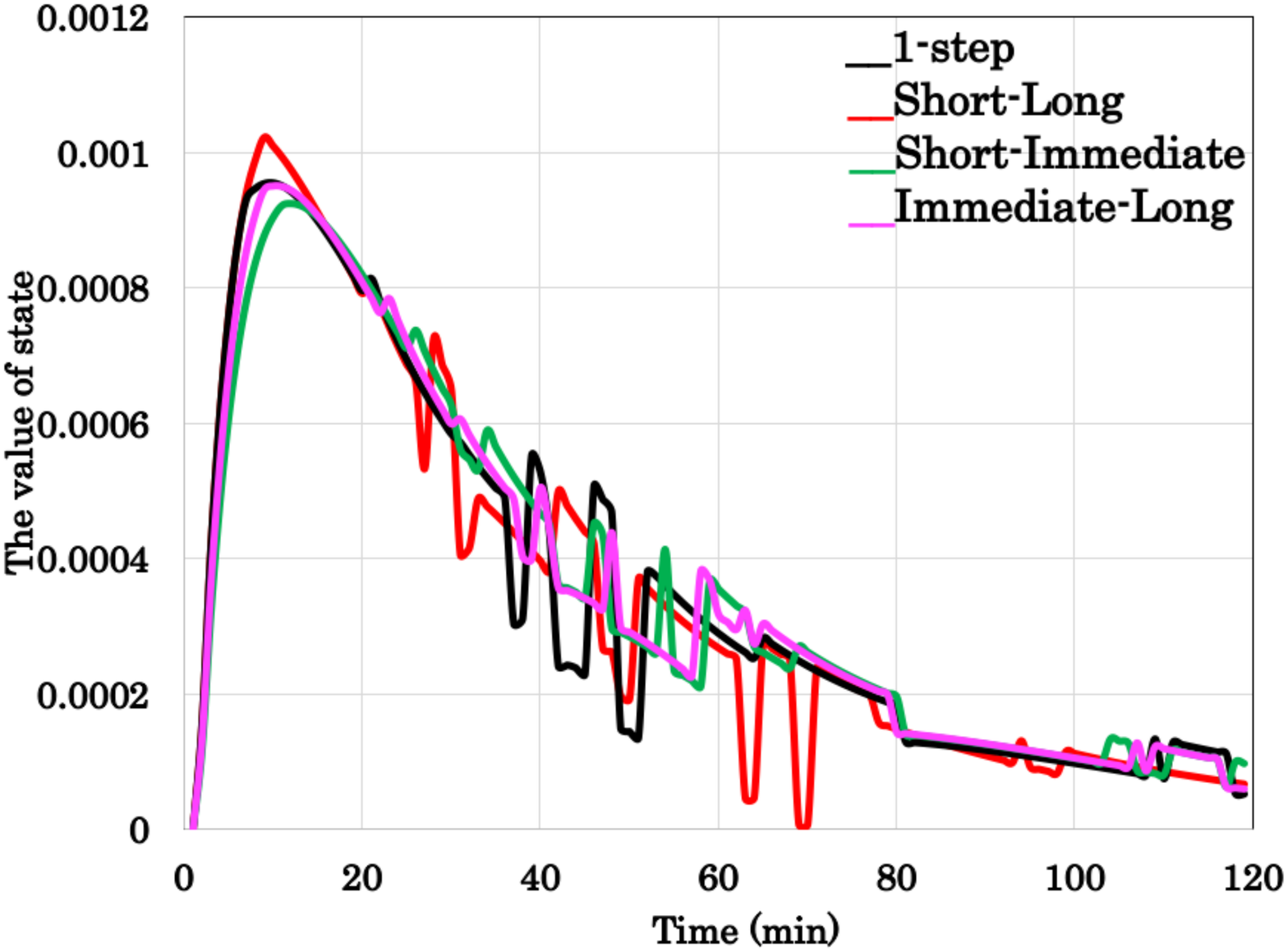}
    \hspace{-12.6ex}
    \caption{The the expected rewards at different steps during the whole reaction time.}
\end{figure}

In this fed-batch process, the differences between $\Delta[C]$ and $\Delta[D]$ (the value of the state) during the previous and the immediate reaction time are greater than that in the latter reaction time. Therefore, more expected benefits can be obtained by the 1-step, short-immediate and immediate-long combination-step compared with the short-long combination-step. However, the control policy of the short-sight and the immediate-sight can emphasize the short and the immediate control results, resulting in a better performance, while the final control results are not the best ones. On the contrary, the long-sight can pay more attention to the final results, and thus generating better final control results. When long-sight is combined with short-sight, the control policy can emphasize both previous and latter control results, and therefore, the control performance of the short-long combination-step is the best. Based on the comparisons with other algorithms, the proposed RVL can achieve the best control results.

\section{Conclusion}
In this paper, we proposed a novel Reinforcement Virtual Learning (RVL) algorithm by creating a virtual space to interact with the agent of RL and the learned virtual policy. The agent of RL can be introduced to learn the real learning policy resulting the feedback to modify the virtual learning policy after interaction with real environment. It is worth noting that the approximated future results of the combinations between short-sight and long-sight through the virtual environment can help the agent to acquire a better real control policy. The proposed RVL overcomes several existing problems, such as uncertain environment, time-variation, and non-linearity. In addition, our experiments demonstrated that the fed-batch process controlled by the proposed RVL can outperform the existing stare-of-the-art algorithms, leading to the effective and stable control performances.

The further work includes applying the proposed RVL to other control applications. For example, RVL can be served for robot control by learning a virtual strategy through a virtual environment of RVL. In addition, this virtual strategy can help robot to achieve some control tasks. The inverse reinforcement learning can replace the basic part of RVL, which can be applied in self-driving as well. When MARL and CNN are applied in both virtual part and basic part, it can tackle some medical issues. The proposed RVL can be combined with graph neural network in some applications as well.

\bibliography{ref}
\bibliographystyle{unsrt}


%





\ifCLASSOPTIONcaptionsoff
  \newpage
\fi

\end{document}